\documentclass{ifacconf}

\usepackage{graphicx}      
\usepackage{natbib}        
\usepackage{enumerate}
\usepackage{amssymb}
\usepackage{amsmath}
\usepackage{algorithm}
\usepackage{algpseudocode}
\usepackage{booktabs}
\usepackage[cmyk]{xcolor}
\begin{document}
\begin{frontmatter}

\title{A Goal-Oriented Reinforcement Learning-Based Path Planning Algorithm for Modular Self-Reconfigurable Satellites} 


\author[1]{Bofei Liu} 
\author[1]{Dong Ye} 
\author[1]{Zunhao Yao}
\author[1]{Zhaowei Sun}

\address[1]{State Key Laboratory of Micro-Spacecraft Rapid Design and Intelligent Cluster, Harbin Institute of Technology, Harbin; 150001, China (e-mail: bofeiliu029@gmail.com; yed@hit.edu.cn; zunhaoyao163@163.com; sunzhaowei@hit.edu.cn)}


\begin{abstract}                
Modular self-reconfigurable satellites refer to satellite clusters composed of individual modular units capable of altering their configurations. The configuration changes enable the execution of diverse tasks and mission objectives. Existing path planning algorithms for reconfiguration often suffer from \emph{high computational complexity}, \emph{poor generalization capability}, and \emph{limited support for diverse target configurations}.
To address these challenges, this paper proposes a \emph{goal-oriented reinforcement learning-based} path planning algorithm. This algorithm is the first to address the challenge that previous reinforcement learning methods failed to overcome, namely handling multiple target configurations. Moreover, techniques such as Hindsight Experience Replay and Invalid Action Masking are incorporated to overcome the significant obstacles posed by sparse rewards and invalid actions. Based on these designs, our model achieves a \emph{95\% and 73\%} success rate in reaching \emph{arbitrary} target configurations in a modular satellite cluster composed of four and six units, respectively.
\end{abstract}


\end{frontmatter}

\section{Introduction}

Modular Self-Reconfigurable Satellites (MSRSs) \citep{MSRS} have emerged as a novel concept in the aerospace field and represent a significant direction for future satellite design. An MSRS is composed of standardized modules, which are relatively small in size and capable of completely altering their configurations through mechanisms such as detaching, docking, and relative motion \citep{modular_motion}. In recent years, with the explosive increase in the number of satellites launched into low Earth orbit and the growing complexity of space missions, organizations such as NASA \citep{NASA} have identified highly standardized, multifunctional, and autonomous MSRSs as one of their key research priorities.


Compared to the traditional single-satellite single-mission paradigm, MSRSs offer notable advantages: they streamline satellite design and manufacturing through standardized modules, enhance deployment efficiency of space systems such as space telescopes \citep{telescope} and orbital solar power stations \citep{station} via batch launches and autonomous on-orbit assembly, and extend operational lifespan of space assets by enabling in-orbit repair, refueling, and recovery \citep{recycle1,recycle2,fuel} , thereby mitigating space debris accumulation.

In recent years, the design of electromagnetic rotating cube modules \citep{magnet} has become relatively mature. These modules eliminate the need for complex mechanical components such as motors, gears, or transmission systems. Moreover, the use of electromagnetic forces enables simultaneous actuation and attachment, eliminating the need for distinct mechanisms for each function. In addition, electromagnetic actuation provides precise force/torque control, enabling accurate docking and detaching operations \citep{force_torque}. As a result, this approach is particularly well suited for space missions due to its simplicity, reliability, and contamination-free nature \citep{magnet_advan}.
This paper focuses primarily on the path planning algorithm for electromagnetic rotating cube configurations.

To date, several studies \citep{A*,K-M} have applied traditional path planning algorithms such as A* and the K-M algorithm to the problem of path planning for electromagnetic rotating cube modules. These methods can achieve successful reconfiguration to target formations with a certain degree of reliability. However, traditional path planning algorithms often suffer from high computational complexity. For instance, the A* algorithm has a time complexity of $O(n^2 \log n)$.
Experimental results \citep{actor-critic} have shown that, A* requires approximately \emph{100 to 200} times more computation time than reinforcement learning approaches. Consequently, traditional path planning algorithms are not well-suited for scenarios where satellites must quickly adapt their configurations to respond to new mission demands.

With the advancement of machine learning technologies, researchers have begun exploring reinforcement learning (RL) algorithms for the path planning of MSRSs. For example, \cite{actor-critic} employed an actor-critic framework to achieve globally observable path planning, while \cite{LSTM} utilized an LSTM-based approach to handle partially observable scenarios for single-module path planning.
Although these approaches can successfully generate feasible paths under specified conditions, they share a common limitation: the target configurations are \emph{fixed} prior to training. As a result, trained models can only handle the single specific goal configuration seen during training and are unable to generalize to more unseen configurations. This means that for a satellite to adapt to different target formations, a separate model must be trained for each specific configuration \textemdash a strategy that is clearly impractical. Even with a relatively small number of modules, the number of potential target configurations is vast, making exhaustive training infeasible.

To address the aforementioned limitations, we adopt a goal-oriented reinforcement learning approach. By incorporating the target configuration as part of the model’s input, our method enables policy learning for all sets of goal configurations.
To improve sample efficiency and mitigate the sparse reward problem, we integrate Hindsight Experience Replay into the replay buffer, which accelerates training and enhances learning from failed trajectories. For invalid action avoidance, we apply action masking techniques, allowing the model not only to learn correct strategies but also to effectively avoid infeasible ones. In addition, we refine the reward function design to allow the agent to extract richer information from its trajectories.

The main contributions of this work are as follows:
\begin{enumerate}[(1)]
    \item The concept of goal-conditioning is introduced into the Markov modeling of the MSRSs path planning problem for the first time, enabling the model to learn distinct strategies for different target configurations.
    \item The Soft Actor-Critic (SAC) algorithm is employed for training. The proposed method achieves high success rates in reconfiguration tasks involving a satellite composed of a limited number of modules.
    \item The reward function is redesigned to provide richer learning signals, and the training process is enhanced by incorporating Hindsight Experience Replay and action masking techniques.
\end{enumerate}
The rest of this article is organized as follows. Section 2 introduces the preliminary background. Section 3 formulates the path planning problem as a Markov Decision Process. Section 4 presents the reinforcement learning algorithm used in our approach. Section 5 describes the experimental setup and report the results. Section 6 concludes the paper and outlines potential directions for future work.

\section{Preliminaries}

\subsection{Goal-Oriented Reinforcement Learning}

Goal-Oriented Reinforcement Learning (GORL) \citep{GORL} is a subfield of reinforcement learning primarily applied in the domain of robotics. Before delving into the details of GORL, we first provide a brief overview of conventional reinforcement learning and the Markov Decision Process (MDP) framework.

The MDP \citep{mdp} is the theoretical foundation of RL, which is denoted as a tuple $(\mathcal{S}, \mathcal{A}, \mathcal{T}, r, \gamma, \rho_0)$, where $\mathcal{S}$, $\mathcal{A}$, $\gamma$ and $\rho_0$ denote the state space, action space, discount factor and the distribution of initial states, respectively.$\mathcal{T}:\mathcal{S} \times \mathcal{A} \times \mathcal{S} \to [0,1]$ is the dynamics transition function, and $r:\mathcal{S} \times \mathcal{A} \to \mathbb{R} $ is the reward function. Reinforcement learning aims to learn policy function $\pi : \mathcal{S} \to \mathcal{A}$ that maximizes expected cumulative return:

\begin{equation} \label{equ}
    J(\pi)=\mathbb{E}_{\substack{a_{t} \sim \pi\left(\cdot \mid s_{t}\right) \\ s_{t+1} \sim \mathcal{T}\left(\cdot \mid s_{t}, a_{t}\right)}}\left[\sum_{t} \gamma^{t} r\left(s_{t}, a_{t}\right)\right] .
\end{equation}

Standard reinforcement learning typically trains an agent to accomplish a single, specific task. However, in our setting, we aim for the agent to adapt its strategy based on varying target configurations, thereby achieving generalization in reconfiguration policies.
To tackle such a problem, the formulation of GORL augments the MDP with an extra tuple $(\mathcal{G}, p_g, \phi)$ as a goal-augmented MDP (GA-MDP), where $\mathcal{G}$ denotes the space of goals describing the tasks, $p_g$ represents the desired goal distribution of the environment, and $\phi:\mathcal{S} \to \mathcal{G}$ is a tractable mapping function that maps the state to a specific goal. In GA-MDP, the reward function $r:\mathcal{S} \times \mathcal{A} \times \mathcal{G} \to \mathbb{R}$ is defined with the goals, and therefore the objective of GORL is to reach goal states via a goal-conditioned policy $\pi : \mathcal{S} \times \mathcal{G} \times \mathcal{A} \to [0,1]$ that maximizes the expectation of the cumulative return over the goal distribution:

\begin{equation}
    J(\pi)=\mathbb{E}_{\substack{a_{t} \sim \pi\left(\cdot \mid s_{t},g\right),g\sim p_g \\ s_{t+1} \sim \mathcal{T}\left(\cdot \mid s_{t}, a_{t}\right)}}\left[\sum_{t} \gamma^{t} r\left(s_{t}, a_{t},g\right)\right] .
\end{equation}

\subsection{Hindsight Experience Replay}

Sparse rewards pose a major challenge in RL, particularly during early training when limited feedback hinders policy development and necessitates extensive exploration, thus prolonging convergence \citep{sparse-reward}. This issue is exacerbated in multi-goal settings , where the agent must learn policies for a wide range of goals. The increased diversity of required behaviors further reduces the likelihood of discovering effective strategies for any specific goal through random exploration. To alleviate this, we adopt the Hindsight Experience Replay (HER) algorithm \citep{HER}.

HER enhances learning efficiency by transforming failed experiences into learning opportunities. Instead of discarding unsuccessful episodes, HER relabels trajectories with alternative goals achieved during exploration. Formally, given a trajectory $\tau = {(s_0, a_0, r_0, s_1), ..., (s_T, a_T, r_T, s_{T+1})}$ generated under a policy $\pi$ for goal $g$, HER samples substitute goals $g' = \phi(s_a)$, where $s_a$ is a future state, and recomputes rewards accordingly. The relabeled experiences $(s_t, a_t, r_t', s_{t+1}, g')$ are then utilized to update the policy, allowing the agent to learn from outcomes it was able to achieve, even if they differ from the original goal.









\subsection{Invalid Action Masking}


During the reconfiguration of MSRSs, actions causing interference or structural disconnection are invalid and must be avoided. We address this by employing invalid action masking technique \citep{mask}, which sets the gradients corresponding to the logits of the invalid action to zero. This enables the model to learn effective policies while avoiding infeasible actions, thereby enhancing the robustness of the learned strategy. Compared to penalty-based or resampling methods, invalid action masking better stabilizes learning and accelerates convergence.

The formulation for invalid action masking is as follows:
\begin{equation}
\begin{aligned}
&\pi(a \mid s, g) = \frac{\exp(\text{mask}(a))}{\sum_{a'} \exp(\text{mask}(a'))}\\
&\text{mask}(a)=\left\{\begin{matrix}
 a      & \text{if a is valid in s}\\
-\infty & \text{otherwise}
\end{matrix}\right.
\end{aligned}
\end{equation}

\section{Problem Formulation}

\begin{figure}[htbp]
\includegraphics[width=8.4cm]{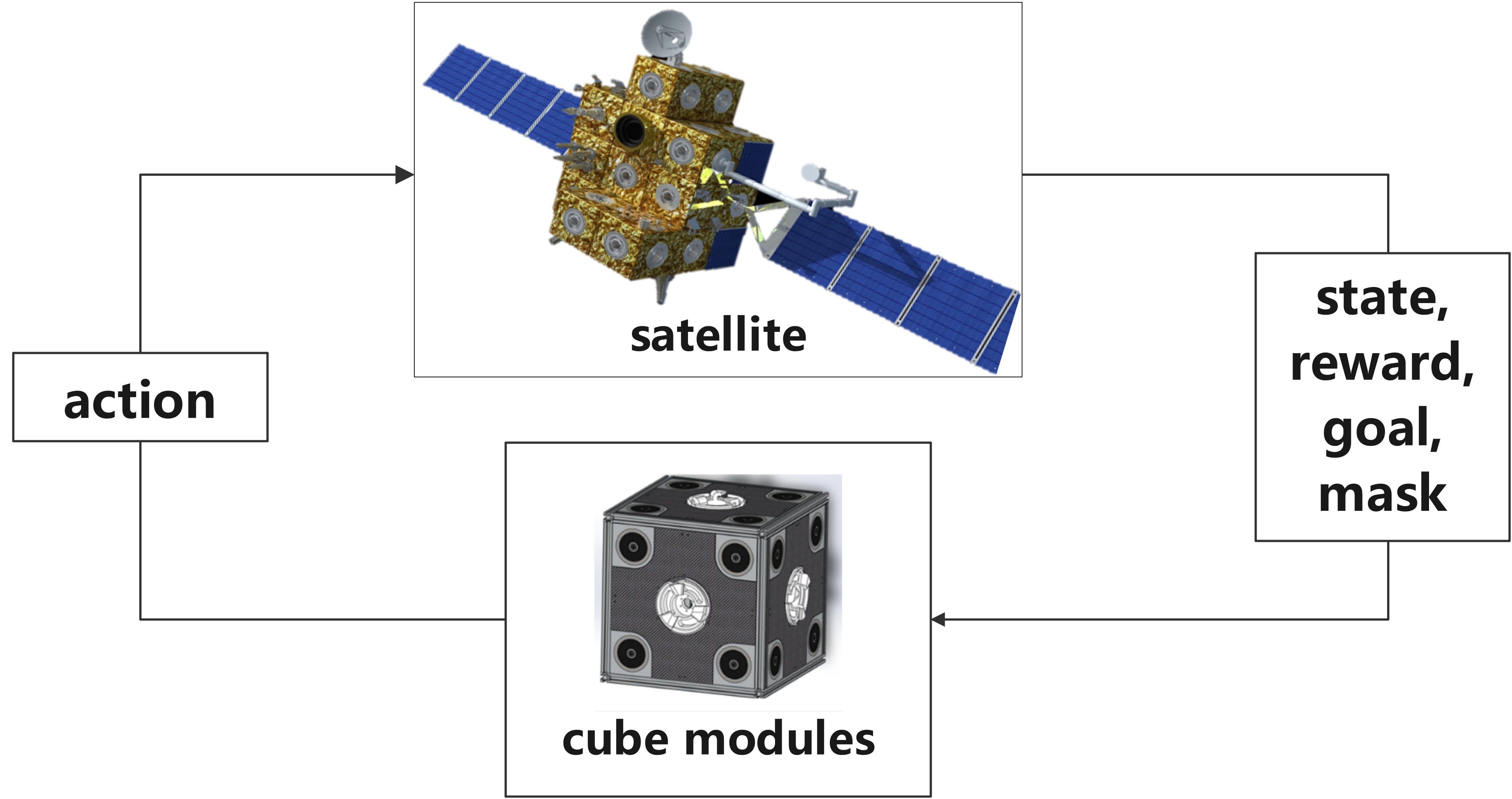}    
\caption{The formulated GA-MDP environment.} 
\label{fig:MDP}
\end{figure}

In this section, we model the path planning problem of electromagnetic rotating cube modules as a GA-MDP , as illustrated in figure~\ref{fig:MDP}.

We provide a detailed description of the model components in the following.
\subsection{State Space and Goal Space}


Module 0 is fixed at $(0, 0, 0)$ during reconfiguration. The state space comprises the 3D coordinates of all modules, while a goal space, defined by the target configuration's module coordinates, is introduced to guide the reconfiguration process.

\subsection{Action Space}

\subsubsection{Module Kinematics Model} \label{sec:kinematics model}

Due to the microgravity and low damping characteristics of the space environment, we model the single-step motion of rotating cube modules using the standard Pivoting Cube Model (PCM). This model characterizes each rotational movement with three key elements: rotation axis, rotation direction and rotation angle. Once these three elements are specified, a unique rotational motion is determined.

\begin{figure}[htbp]
\includegraphics[width=8.4cm]{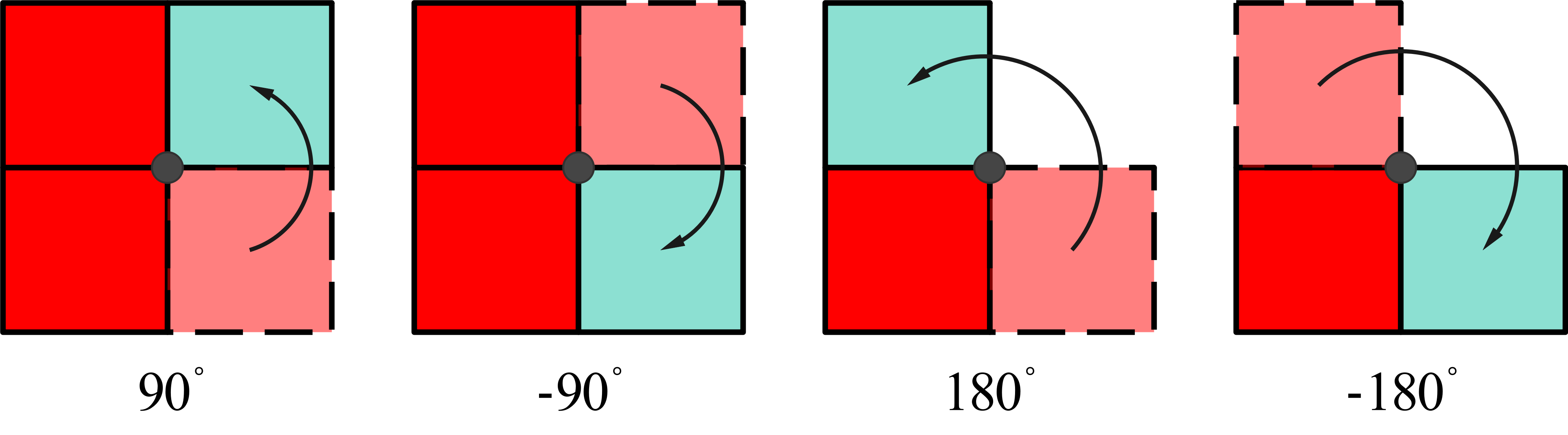}    
\caption{Illustration of module rotation.} 
\label{fig:rotate}
\end{figure}

For a given rotation axis, a cube module in the satellite can perform four possible actions: a $90^\circ$ rotation, a $-90^\circ$ rotation, a $180^\circ$ rotation and a $-180^\circ$ rotation , as illustrated in figure~\ref{fig:rotate}. 
Here, the rotation direction is defined such that a counterclockwise rotation, when viewed from the positive direction of the rotation axis, is considered positive.

\subsubsection{Action Space Definition}

According to the module kinematic model described in Section~\ref{sec:kinematics model}, a single module can perform 48 distinct actions in total. These 48 actions are encoded as actions 1 through 48. In addition, we define action 0 as a no-op, representing the choice of taking no action.
Since the path planning algorithm proposed in this work follows a global scheme in which only one module is allowed to move at each time step, the final action space $\mathcal{A}$ is defined as $\left \{a^* \times (i+1)\right \} \cup \left \{ 0 \right \}$, where $ a^{*} \in \{1, 2, \dots, 48\} $ denotes the action index, and $ i \in \{0, 1, 2, \dots\} $ is the index of the module.

\subsubsection{Invalid Action Definition}

Here, we briefly introduce the definition of invalid actions, which fall into the following three categories:

\begin{itemize}
    \item Type A: the module attempts to rotate around an edge that is not connected to a neighboring module.
    \item Type B: the module's movement trajectory results in interference with other modules in the structure.
    \item Type C: after the action is executed, the resulting configuration becomes disconnected\footnote{The judgment of configuration connectivity is abstracted as a graph theory problem, and in this work, we adopt the \emph{Warshall algorithm} to determine connectivity.}.
\end{itemize}

\begin{figure}[htbp]
\includegraphics[width=8.4cm]{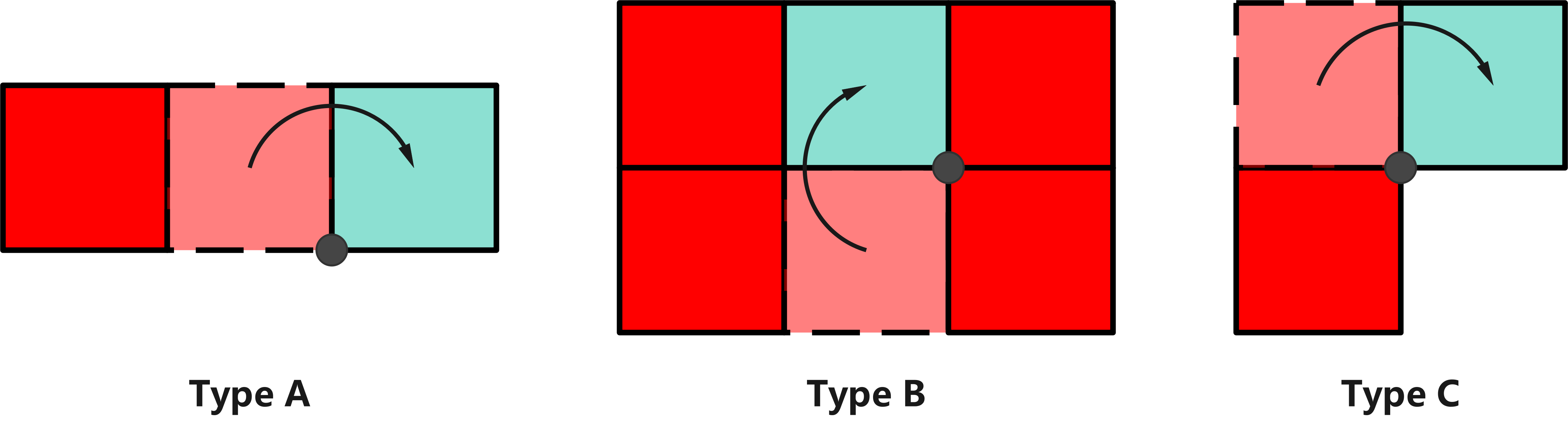}    
\caption{Examples of invalid action.} 
\label{fig:invalid}
\end{figure}

Examples of invalid actions are illustrated in figure~\ref{fig:invalid}.

\subsection{Reward Function}

\subsubsection{Configuration Distance Computation}

To simplify the problem, we assume that all modules are identical. The distance between the current and target configurations is defined as the minimum total Euclidean distance achieved through optimal matching, rather than by simply summing distances between fixed module pairs.
We adopt the \emph{Hungarian algorithm} to find the optimal matching between modules in the current and target configurations. Based on this matching, we define the configuration distance $ \mathcal{D}(\mathcal{S}, \mathcal{G}) $ between the current state space $ \mathcal{S} $ and the goal space $ \mathcal{G} $ as follows:

\begin{equation}
    \mathcal{D}(\mathcal{S}, \mathcal{G}) = \min_{\substack{\sigma \in \mathcal{M}\\ s \in \mathcal{S}\\ g \sim \mathcal{G}}}\sum_{i=1}^{n} ||s_i-g_{\sigma (i)}||_2
\end{equation}

,where $\mathcal{M}$ denotes matches of all possible permutations , $g$ denotes the target position of each module under the corresponding assignment $\sigma$ , $n$ denotes the total number of modules.

\subsubsection{Reward Function Design}

A well-designed reward function is crucial in reinforcement learning, which effectively guides policy optimization by allowing agents to reasonably evaluate their actions.

Previous work on path planning for MSRSs has often relied on overly simplistic reward functions, such as \cite{actor-critic}, which assign a reward of 0 when the satellite's configuration matches the target, and a penalty of -1 otherwise. This approach suffers from \emph{sparse reward problem}, as frequent -1 rewards during the early stages of exploration offer minimal feedback, hindering effective policy learning and slowing training.

To address the aforementioned issue, we carefully designed the following reward function through empirical analysis:

\begin{equation}
    r=\left\{\begin{matrix}
 \mathcal{D}(\mathcal{S'},\mathcal{G}) - \mathcal{D}(\mathcal{S},\mathcal{G}) & \text{not done}\\ 
  +10 & \text{done}\\
  -1 & \text{no-op before done}
\end{matrix}\right.
\end{equation}

In the above function, $\mathcal{S}$ and $\mathcal{S'}$ represent the state space of the configuration before and after the action, respectively. 
With this design, the agent can effectively evaluate whether its current action moves the module closer to the target configuration (positive reward) or further away (negative reward). 
In addition, we assign a reward of -1 when the agent chooses a ``no-op" (no movement) action before reaching the target. This discourages the agent from falling into suboptimal strategies where it remains stationary without completing the reconfiguration.

\section{ALGORITHM}

This section offers a comprehensive description of the reinforcement learning algorithms employed in our study.

\subsection{Soft Actor-Critic}
Soft Actor-Critic (SAC) \citep{SAC} is a deep reinforcement learning algorithm based on the maximum entropy framework. It adopts an actor-critic architecture to maximize cumulative reward while encouraging exploration by maximizing policy entropy, helping prevent premature convergence to suboptimal solutions.
Compared to the standard objective in (\ref{equ}), the cumulative reward that SAC aims to maximize is modified as follows:

\begin{equation}
    J(\pi)=\sum_{t} \mathbb{E}_{(s_{t},a_{t}) \sim \tau_{\pi}} \left[ \gamma^{t} (r\left(s_{t}, a_{t}\right)
    +\alpha \mathcal{H}(\pi(\cdot|s_t))) \right] .
\end{equation}

, where $\tau_{\pi}$ is the state-action distribution under policy $\pi$ , $\mathcal{H}(\pi(\cdot|s_t)) = -\mathbb{E}_{a_t \sim \pi}[\log\pi(a_t|s_t)]$ is the entropy of the policy at state $s_t$ , $\alpha$ is the temperature coefficient that balances the trade-off between exploration and exploitation.

In addition, SAC incorporates several other key techniques to enhance stability and performance:

\begin{itemize}
    \item Double Q-network architecture: mitigate the overestimation bias in value function approximation.
    \item Soft update for target networks: utilize a Polyak averaging strategy to ensure smoother convergence.
    \item Automatic entropy tuning: balance the trade-off between exploration and exploitation.
\end{itemize}


\subsection{SAC for Discrete Action in GORL}

In our work, since the rotation of modules corresponds to a discrete action space, we adopt a discrete action variant of the SAC algorithm (SAC-Discrete) \citep{DSAC} to address the path planning problem of MSRSs. Building upon the base algorithm, we incorporate several enhancements to address specific challenges in our scenario:

\begin{itemize}
    \item Goal space integration: We introduce goal space as part of the input, enabling the agent to learn generalized policies conditioned on varying target configurations, thereby addressing the generalization issue.
    \item Hindsight Experience Replay: To mitigate the sparse reward problem inherent in multi-goal reinforcement learning tasks, we utilize HER to augment the training data by treating failed episodes as successful ones for alternate goals.
    \item Action masking: To ensure safety and physical feasibility, we implement invalid action masking, which prevents the agent from selecting illegal actions such as those leading to collisions or disconnected configurations.
\end{itemize}

\begin{figure}[htbp]
\includegraphics[width=8.4cm]{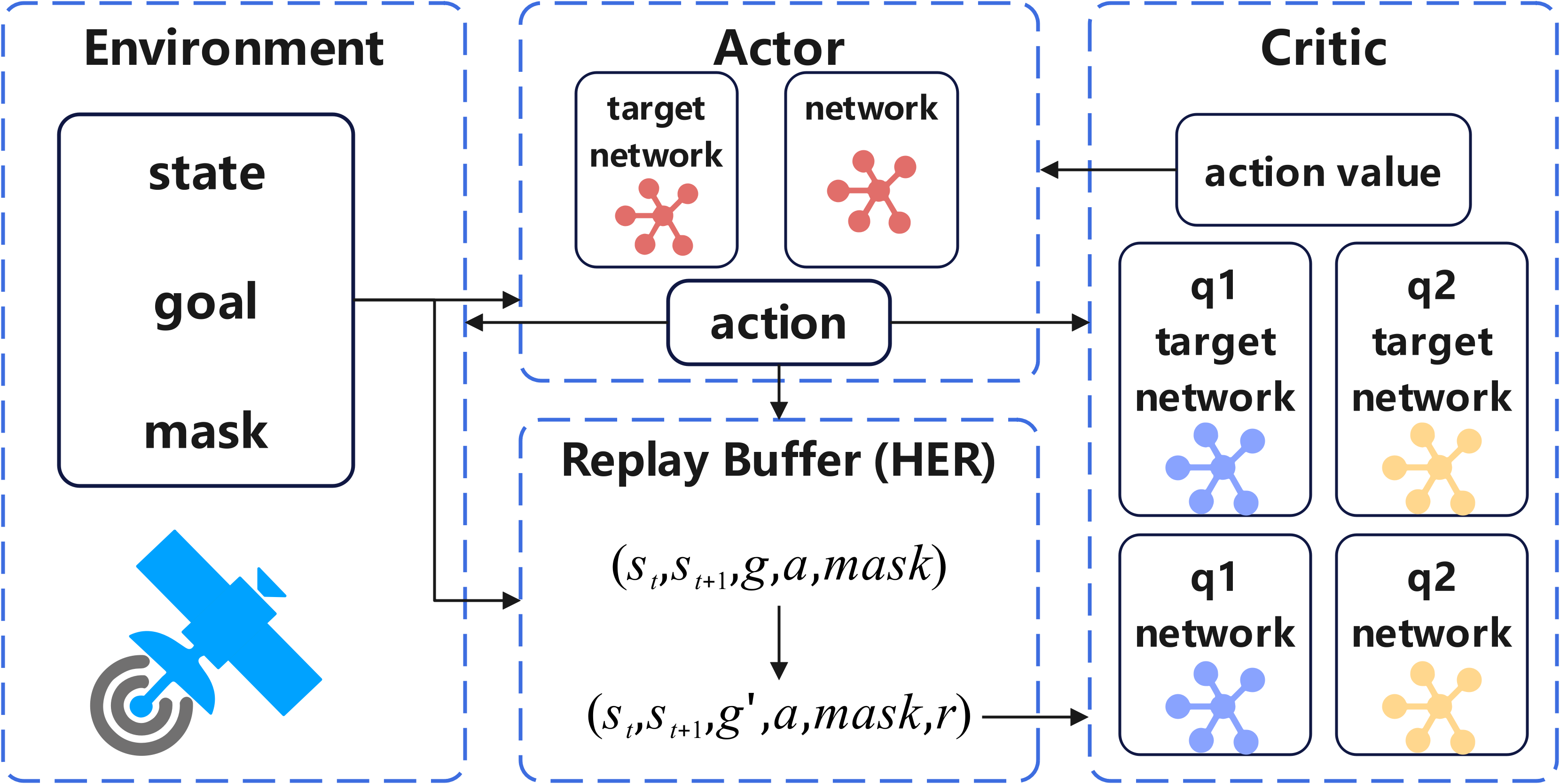}    
\caption{The proposed learning framework for our work.} 
\label{fig:frame}
\end{figure}

The overall model framework is illustrated in figure~\ref{fig:frame}.

Distinct from the continuous-action SAC algorithm, below we present the detailed formulation and mechanisms of the SAC-Discrete algorithm adapted to Goal-Oriented Reinforcement Learning.

The soft Q-function is updated by minimizing the following loss:

\begin{equation}\label{equ:Q}
\mathcal{L}_Q = \mathbb{E}_{(s,a,r,s',g)} \left[ \left( Q(s,a,g;\phi) - \hat{Q}(s,a,g;\phi) \right)^2 \right]
\end{equation}

with the target:

\begin{equation}
\begin{array}{ll}
\hat{Q}(s,a,g) = &r + \gamma \mathbb{E}_{a' \sim \pi(\cdot \mid s', g; \theta)} [ Q_{\text{target}}(s',a',g;\phi)\\
&- \alpha \log \pi(a'|s', g;\theta)]
\end{array}
\end{equation}

where $Q_{\text{target}}$ is the target Q-network.

The policy is optimized by minimizing:

\begin{equation}\label{equ:actor}
\mathcal{L}_\pi = \mathbb{E}_{s \sim \mathcal{D},\, a \sim \pi(\cdot \mid s, g;\theta)} \left[ \alpha \log \pi(a | s, g;\theta) - Q(s,a,g;\phi) \right]
\end{equation}

where $\mathcal{D}$ is a replay buffer of past experiences.

The entropy temperature $\alpha$ is adjusted automatically by minimizing:

\begin{equation}\label{equ:alpha}
\mathcal{L}_\alpha = \mathbb{E}_{a \sim \pi(\cdot \mid s, g;\theta)} \left[ -\alpha \left( \log \pi(a | s, g;\theta) + \mathcal{H}_{\text{target}} \right) \right]
\end{equation}

where $\mathcal{H}_{\text{target}}$ is the target entropy of policy.

Combining all these changes, our algorithm for SAC with discrete actions in GORL is given by Algorithm~\ref{alg3}.

\begin{algorithm}[htbp]
\caption{SAC-Discrete in GORL for our work}
\label{alg3}
\begin{algorithmic}
    \State Initialize actor network with $\theta$, critic networks with $\phi$
    \State Initialize target networks: $\phi_1' \gets \phi_1$, $\phi_2' \gets \phi_2$
    \State Initialize temperature factor $\alpha$, replay buffer $\mathcal{D}$
    \For{each iteration}
        \For{each environment step}
            \State $a_t \sim \pi(\cdot \mid s_t, g; \theta)$
            \State $s_{t+1}, g, \text{mask} \sim p(\cdot \mid s_t, a_t)$
            \State $\mathcal{D} \gets \mathcal{D} \cup \{(s_t, a_t, s_{t+1}, g, \text{mask})\}$
        \EndFor
        \For{each gradient step}
            \State Apply HER algorithm
            \State Update critic networks using \eqref{equ:Q}
            \If{$t \bmod 2 = 0$}
                \State Update actor network using \eqref{equ:actor}
                \State Update $\alpha$ using \eqref{equ:alpha}
                \If{update target network cycle}
                    \State $\phi_1' \gets \tau \phi_1 + (1 - \tau)\phi_1'$
                    \State $\phi_2' \gets \tau \phi_2 + (1 - \tau)\phi_2'$
                \EndIf
            \EndIf
        \EndFor
    \EndFor
\end{algorithmic}
\end{algorithm}

\section{Experiments}

Based on the aforementioned content, we conducted training and testing of self-reconfiguration path planning for satellites consisting of 4 and 6 modules, respectively. Table~\ref{table:common_hyperparameters} presents a subset of the hyperparameter settings used during the training process.

\begin{table}[htbp]
\centering
\caption{Hyperparameter used in our train.}
\label{table:common_hyperparameters}
\begin{tabular}{c|c|c}
\toprule
Hyper-parameters & 4 modules & 6 modules \\
\midrule
Epoch number & 500 & 1200 \\ 
Batch number & 200 & 500 \\ 
Batch size & 512 & 256   \\
Buffer size &  2e7  &  1e6 \\
actor lr & 9e-4  &  9e-4  \\
critic lr & 9e-4  &  9e-4  \\
$\alpha$ lr & 3e-4  & 3e-4  \\
Initial value of $\log \alpha$& -2.0 & -1.0 \\
$\tau$ & 0.005 & 0.005 \\
Discount factor & 0.98 & 0.98 \\
\bottomrule
\end{tabular}
\end{table}

\begin{figure}[htbp]
\includegraphics[width=8.4cm]{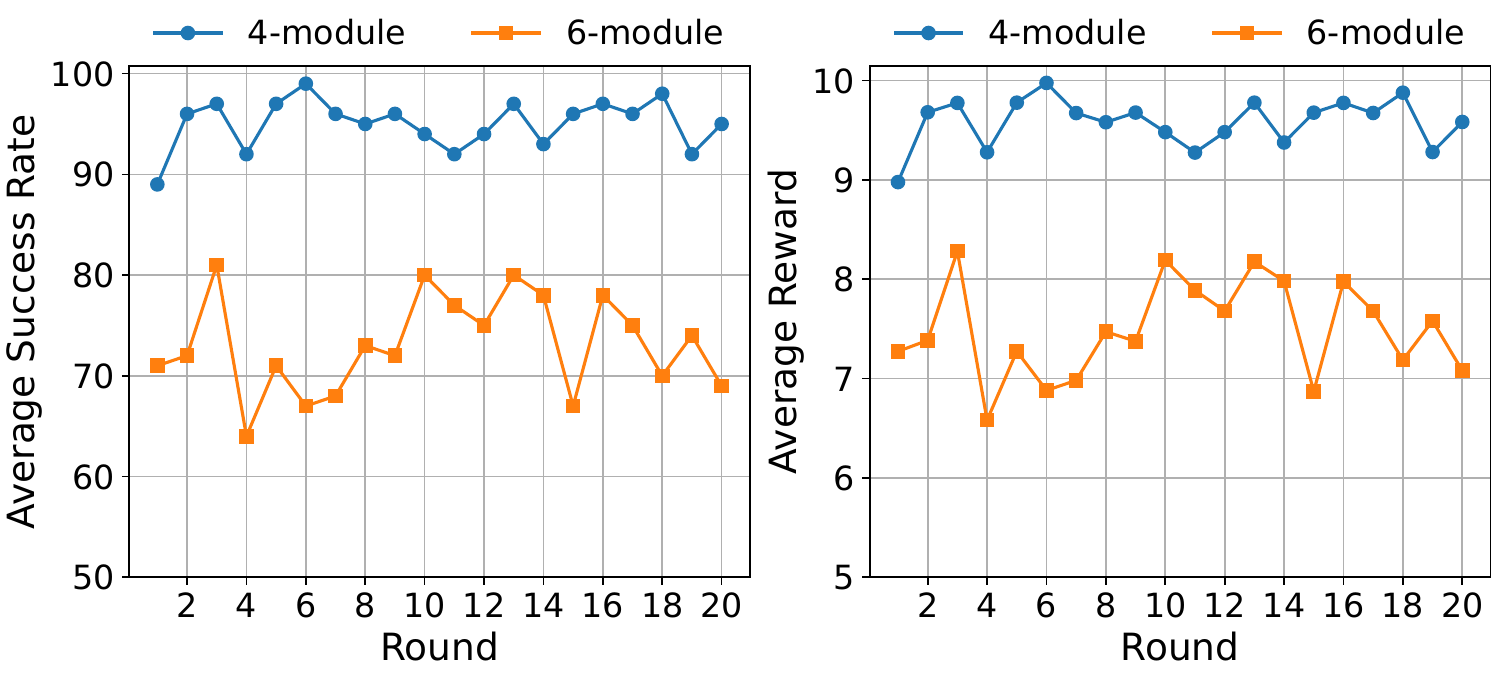}    
\caption{Average success rate and average reward.} 
\label{fig:compare}
\end{figure}

Figure~\ref{fig:compare} presents the average success rate and average reward of the reconfiguration path planning model for 4-module and 6-module satellites. Each data point is obtained by averaging the results over 100 randomly generated target configurations, and a total of 20 rounds of testing are conducted.

The experimental results demonstrate that our model can achieve a high reconfiguration success rate for \emph{arbitrary} target configurations after training. Specifically, the overall average success rate for the 4-module configuration is 95.05\% , while 73.1\% for the 6-module configuration.

\begin{figure}[htbp]
\includegraphics[width=8.4cm]{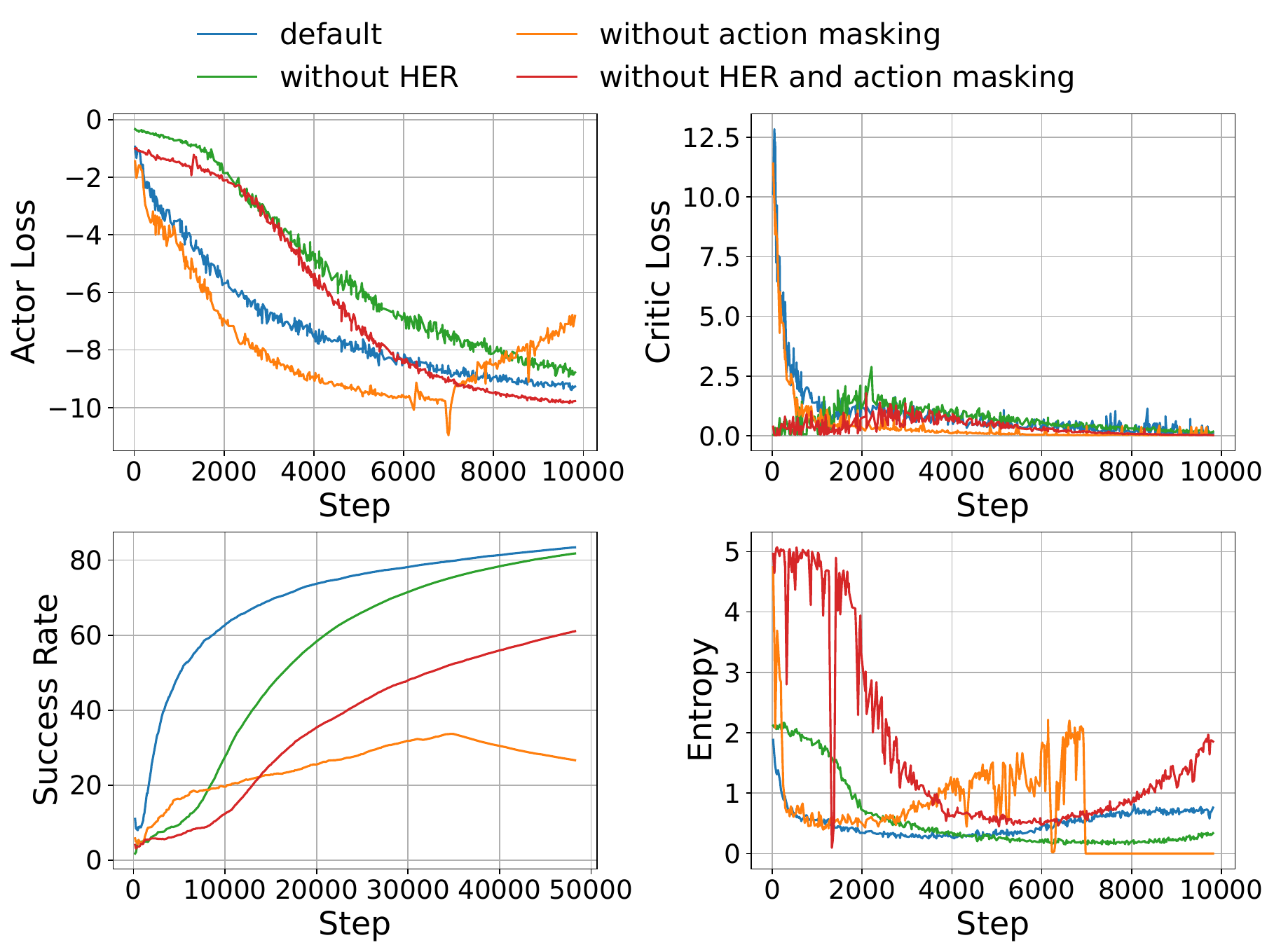}    
\caption{Results of ablation studies in the early stages of training.} 
\label{fig:4}
\end{figure}

In addition, we conducted ablation studies to evaluate the contributions of the HER and action masking techniques in our experiments. Using the 4-module configuration as the test case, we plotted the curves of actor loss, critic loss, success rate and policy entropy during the training process, as shown in figure~\ref{fig:4}.

\begin{figure}[htbp]
\includegraphics[width=8.4cm]{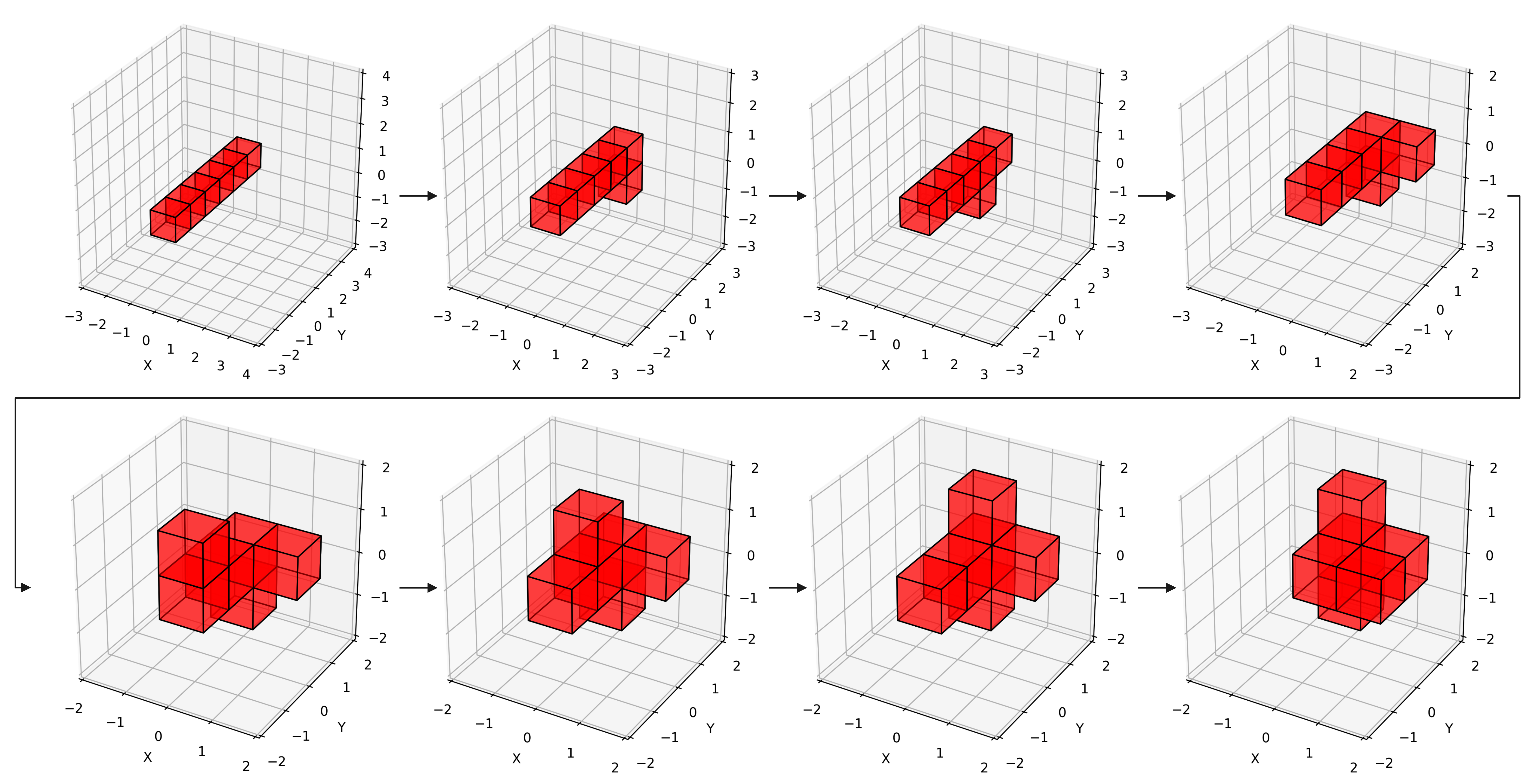}    
\caption{A reconfiguration example of our algorithm.} 
\label{fig:example}
\end{figure}

The experimental results demonstrate that, compared to models without HER or action masking, our method is able to identify optimal strategies more efficiently and avoids policy entropy divergence caused by convergence to suboptimal strategies. Figure~\ref{fig:example} illustrates an example of the reconfiguration produced by our algorithm. The code of the proposed approach and the learned models are available at https://github.com/perfactliu/GORL4MSRS.

\section{Conclusion}


This paper focuses on solving the path planning problem of electromagnetic rotating modular satellites using reinforcement learning algorithms, and introduces the concept of goals into the Markov modeling of this problem for the first time. To improve the training stability and efficiency, we incorporate several mechanisms into the learning framework, including action masking , Hindsight Experience Replay , and a carefully designed reward function that provides more informative feedback during training. Experimental results demonstrate that, for satellites composed of four and six modules, our model achieves a average success rate of 95\% and 73\% respectively across arbitrary goal configurations.

The proposed model is capable of solving the path planning problem for any target configuration of electromagnetic rotating MSRSs, effectively addressing the generalization limitation of previous studies. However, the current approach still suffers from high training time costs, especially when tackling larger modular satellite systems. Future work may explore two potential directions for improvement: (1) utilizing $A^*$ algorithm to pre-sample trajectory , and (2) considering applying distributed action modeling methods to reduce the dimensionality of the action space.

\begin{ack}
National Key R\&D Program of China(2021YFC2202900), the National Natural Science Foundation of China(62203145), the China Postdoctoral Science Foundation(2022M710948), the Civil Aerospace Research Project of China(D030312).
\end{ack}

\bibliography{ifacconf}             
                                                   







\end{document}